\documentclass[letterpaper, 10 pt, journal, twoside]{IEEEtran}

\IEEEoverridecommandlockouts                              % This command is only needed if 
\usepackage{graphics} % for pdf, bitmapped graphics files
\usepackage{epsfig} % for postscript graphics files
\usepackage{mathptmx} % assumes new font selection scheme installed
\usepackage{times} % assumes new font selection scheme installed
\usepackage{amsmath} % assumes amsmath package installed
\usepackage{amssymb}  % assumes amsmath package installed

\usepackage{algorithmic}
\graphicspath{ {Images/} }
\usepackage{url}

\usepackage{breakurl}
\usepackage[breaklinks]{hyperref}
\usepackage{array}
\usepackage{graphicx}
\usepackage{multirow}
\usepackage{tabu}
\usepackage{cleveref}

\usepackage{xcolor}

\title{Benchmarking Simulated Robotic Manipulation\\ through a Real World Dataset}

\author{Jack Collins$^{1,2}$, Jessie McVicar$^{1,2}$, David Wedlock$^{1,2}$, Ross Brown$^{2}$, David Howard$^{1}$ and J\"urgen Leitner$^{2,3}$%
\thanks{Manuscript received: August, 16, 2019; Accepted November, 1, 2019.}%Use only for final RAL version
\thanks{This paper was recommended for publication by Editor Han Ding upon evaluation of the Associate Editor and Reviewers' comments.
This work was supported by a Data61 PhD Scholarship.} %Use only for final RAL version
\thanks{$^{1}$Jack Collins, Jessie McVicar, David Wedlock and David Howard are with Data61, Commonwealth Scientific and Industrial Research Organisation (CSIRO), Brisbane, Australia}%
\thanks{$^{2} $Jack Collins, Jessie McVicar, David Wedlock, Ross Brown and J\"urgen Leitner are with Queensland University of Technology (QUT), Brisbane, Australia}%
\thanks{$^{3}$ J\"urgen Leitner is with the Australian Centre for Robotic Vision (ACRV)}%
\thanks{Digital Object Identifier (DOI): see top of this page.}
}
\begin{document}

\maketitle
% \thispagestyle{empty}
% \pagestyle{empty}

%%%%%%%%%%%%%%%%%%%%%%%%%%%%%%%%%%%%%%%%%%%%%%%%%%%%%%%%%%%%%%%%%%%%%%%%%%%%%%%%
\begin{abstract}
We present a benchmark to facilitate simulated manipulation; an attempt to overcome the obstacles of physical benchmarks through the distribution of a real world, ground truth dataset. Users are given various simulated manipulation tasks with assigned protocols having the objective of replicating the real world results of a recorded dataset. The benchmark comprises of a range of metrics used to characterise the successes of submitted environments whilst providing insight into their deficiencies. 

We apply our benchmark to two simulation environments, PyBullet and V-Rep, and publish the results. All materials required to benchmark an environment, including protocols and the dataset, can be found at the benchmarks' website\footnote{\url{https://research.csiro.au/robotics/manipulation-benchmark/}}.

\end{abstract}

%%%%%%%%%%%%%%%%%%%%%%%%%%%%%%%%%%%%%%%%%%%%%%%%%%%%%%%%%%%%%%%%%%%%%%%%%%%%%%%%

% Keywords appear just beneath the abstract. Use only for final RAL version. 
\begin{IEEEkeywords}
Performance Evaluation and Benchmarking; Simulation and Animation; Grasping; Force and Tactile Sensing
% Force and Tactile Sensing; Performance Evaluation and Benchmarking; Probability and Statistical Methods; Simulation and Animation; Contact Modeling; Dexterous Manipulation; Grasping
\end{IEEEkeywords}

%%%%%%%%%%%%%%%%%%%%%%%%%%%%%%%%%%%%%%%%%%%%%%%%%%%%%%%%%%%%%%%%%%%%%%%%%%%%%%%%
\section{Introduction}

%Less specific, talk about benchmarking in manipulation
% Drop letter for first word of the Introduction
\IEEEPARstart{B}{enchmarks} and datasets are important for research and scientific progress; they foster advancement in key research areas through competition and quantifiable, reproducible results. Robotic manipulation currently lacks such benchmarks that are widely accepted in fields of comparable size and importance, such as Simultaneous Localisation and Mapping (SLAM) \cite{Sturm2012ASystems, Handa2014ASLAM} and object detection in computer vision \cite{Geiger2012AreSuite, Lin2014MicrosoftContext}. Previously-proposed manipulation benchmarks and challenges require access to expensive platforms and specialised environments (i.e.~FetchIt, Amazon Picking Challenge, etc.) with unique objects to be able to assess the capabilities of a system, which limits uptake in the research community at large. This is in contrast to other (robotics) research areas, where employing widely accepted benchmarks -- often comprised of real-world ground truth data -- enables the comparison of different solutions without the need for duplicating the physical hardware. 

%More specific
Manipulation research is often carried out initially or primarily in simulated environments. Simulation is preferable as it (i) facilitates reproducible results, (ii) grants access to potentially unavailable platforms, (iii) is inherently safe, (iv) does not wear or damage physical systems, and (v) can run faster than real-time in addition to running many instances in parallel. However, the deficiencies of physics simulations are easily observed with controllers generated in simulation often proving to be unreliable when transferred to the real world. The phenomena of the reality gap is well known but little has been done to \textit{quantify} this gap~\cite{Collins2019QuantifyingTasks}, both for future improvement of simulators, and to inform the research community. 

%Motivation
The absence of a widely applicable manipulation benchmark that requires no physical hardware and comprises of real-world data is the motivation for our benchmark. Our benchmark and affiliated dataset is positioned to assist researchers and developers in quantifying the reality gap, specifically for physical interaction tasks with robotic arms, leading to advances in simulation-to-reality (sim2real) transfer, as well as physics engines, simulators, and their parameterisations. The need for the benchmark to be sufficiently general to be relevant to many manipulation fields yet specific enough to provide useful information on the system is satisfied through the employment of simulation and quantitative metrics.

% There are a large number of simulation environments that are applicable to simulating robots and manipulation although not all are equal. It has been proven that the basic kinematic movements of manipulators in simulation has been solved yet \cite{me} dynamic interactions with one or more objects is still a challenging task to accurately model.

\begin{figure}[t]
	\centering
	\includegraphics[width=\linewidth]{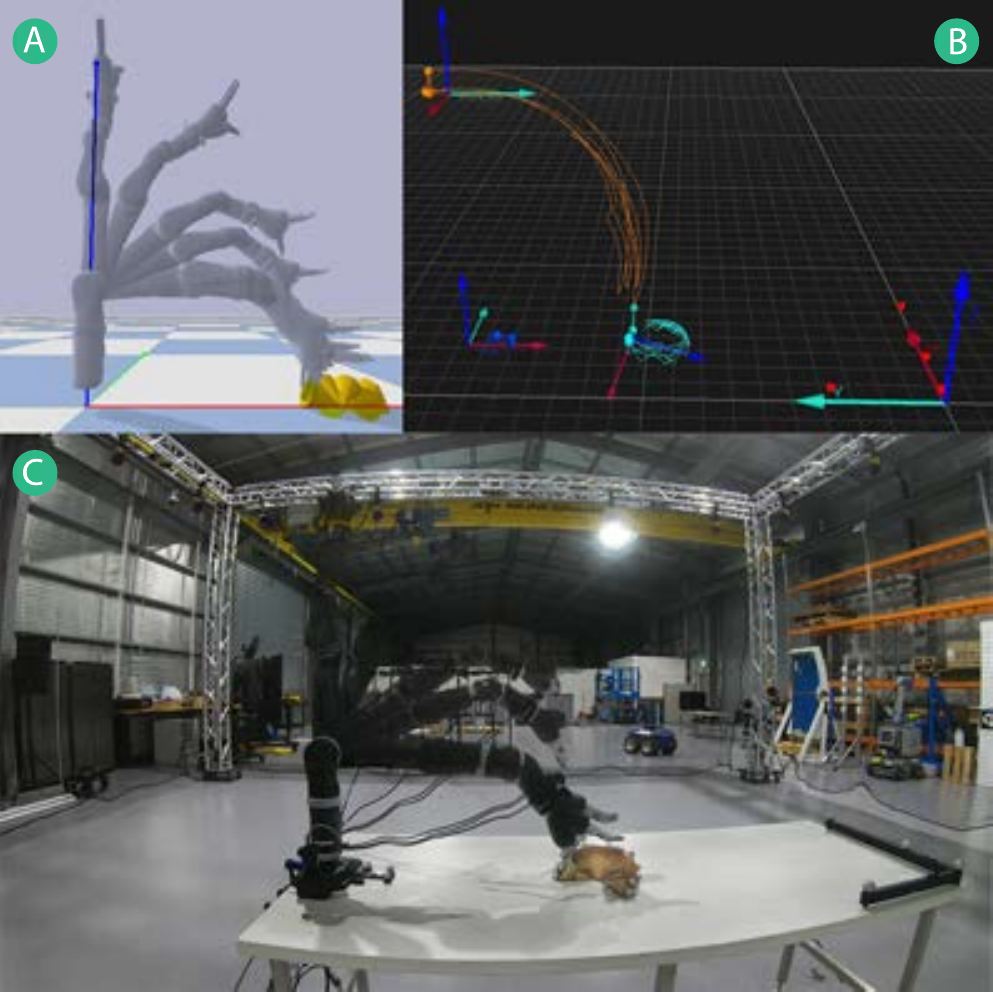}
	\caption{Snapshots through time of the Kinova robotic manipulator performing an interaction task with a cone. (A) Rendering of the PyBullet simulator, (B) recording of the Motion Capture environment, and (C) video of the real world task.}
	\label{VisualAbstract}
	\vspace{-4mm}
\end{figure}

%Our Proposal
 
% \juxi{repeats the previous thing} It is anticipated that this benchmark will allow users to validate their current research environments and allow for comparison between simulation environments promoting robust simulated solutions that transfer well to real-world platforms.
%\juxi{do we need that, says nothing new, and you have not talked about submissions yet...}
%In turn this will allow the shortfalls of submitted solutions to be quantified and better understood.

%A brief summary of what we did
% \juxi{make that the contributions section that Ross mentioned!}
The main contributions of our benchmark are (i) a procedure for comparing simulations to real world recordings, (ii) a dataset of ground truth labelled manipulation tasks performed by a robotic manipulator recorded using a highly accurate motion capture system, and (iii) a chosen subset of metrics for characterising the success of a simulator at reproducing reality (Fig.~\ref{VisualAbstract}). %The dataset of manipulation tasks is recorded using a highly accurate motion capture system and acts as a ground truth.
We expect to expand this dataset over time to cover more tasks and use additional robotic manipulators.

Instructions to replicate each of the tasks in simulation are available in a common format with information on the setup as well as control of the robot. The metrics for comparing simulation performance to the dataset are applied to the results from two common robotic simulators, PyBullet and V-Rep. The exemplar performance is intended as a minimal solution to establish the importance of the chosen metrics, we expect users to improve upon the reported results. 

% Physics engines and robotic simulators have a myriad of user definable parameters to tune that can affect the realism of the simulation environment.  
% Why is this benchmark important?

% A benchmark for simulated manipulation - towards overcoming the obstacles of simulated solutions. We present a dataset and benchmark for researchers and developers working on advancing manipulation, sim2real, physics engines and simulators. 

% Validate your simulation environment against the real world. The closer your simulation replicates the real world the greater the chance that your controller will operate with similar performance on the physical manipulator.

% Why working in simulation is important - reproducibility for groups with limited resource

% Insight into simulators, i.e. importance of parameters, how to tune precision, 

%%%%%%%%%%%%%%%%%%%%%%%%%%%%%%%%%%%%%%%%%%%%%%%%%%%%%%%%%%%%%%%%%%%%%%%%%%%%%%%%
\section{Related Work}

Robotic manipulation is a wide and varied field of research, which poses a challenge when constructing a benchmark that is specific enough to provide useful performance indicators, yet diverse enough to be widely used and accepted. Past benchmarks for real-world manipulation have assessed the mechanical performance of grippers (i.e. grip strength, gripper volume, etc.) \cite{Falco2015GraspingMethodologies, Backus2016AnJoints}, motion and grasp planning \cite{Cohen2013Single-Search}, cooperative or collaborative manipulation \cite{Iocchi2015RoboCupHome}, and challenging specific sub-tasks (i.e. shelf picking, peg-in-hole, deformable bodies, lego assembly, etc.) \cite{Leitner2017TheResearch,Wyk2018ComparativeHand,Lazher2014ModelingGrasping, Yokokohji2000ToySystems}. Few simulated benchmarks exist for manipulation, and those available have been developed to assess the performance of planners and grasping \cite{Ulbrich2011TheManipulation, Popovic2011GraspingUnderstanding, Kootstra2012VisGraB:Grasping}. In contrast to these, we assess the performance of simulators in accurately modelling the real world interaction and motion of a robotic manipulator.

Validation of simulation environments is conducted through the assessment of the underlying physics engine. Commonly, test scenes are developed that are designed to assess elements of the physics engine that are likely to deviate from the real world. Such tests evaluate the numerical integration, constraint stability, collision detection and material properties of the physics engine \cite{Roennau2013EvaluationRobots, Boeing2007EvaluationSystems, Chung2016PredictableEngines}. Tests are evaluated with reference to the real world behaviour by drawing comparisons from the human expectation of the system, or from the derivation of behaviour from first principles. Recently, a new method was introduced that uses a dataset of ground-truth movements to validate these tests \cite{Collins2019QuantifyingTasks}. This new approach assesses the performance of physics engines using complex environments with a robot represented as a collection of rigid bodies controlled using simulated actuators.

Datasets in robotics have become increasingly common due to the wide acceptance of machine learning which frequently requires large amounts of labelled data to train. The manipulation field lacks ground truth datasets that have been available to other fields for some time. Fields that have had access to datasets with ground truth labels (i.e. SLAM and computer vision) have experienced great uptake and utilisation leading to advances in their respective fields. The  most suitable ground-truthing mechanism for robotic manipulation tasks is a Motion Capture system.

Motion capture provides measurements on the six-dimensional pose of rigid-bodies throughout time. Datasets that have been collected through the use of motion capture include human pose estimation \cite{Ionescu2014Human3.6M:Environments}, interactive manipulation performed by a human \cite{Huang2019AManipulation} and sequences of labelled RGB-D images for SLAM \cite{Sturm2012ASystems}. More specifically for tasks by robot manipulators, there are motion capture datasets of robots performing planar pushing and datasets of rigid body contacts, but here the emphasis is on tracking the effects of the robot in the context of a task, not the robot itself \cite{Yu2016MorePushing, Fazeli2017FundamentalModels}. Tracking of a dexterous manipulator to overcome the reality gap is similar to what we propose, however this information isn't available as a re-usable benchmark dataset \cite{Andrychowicz2018LearningManipulation}.

\section{The Benchmark}

Our manipulation benchmark consists of three components:
\begin{enumerate}
    \item A real world dataset of motion capture tasks;
    \item A definition of the tasks to be simulated in the chosen simulation environment; and
    \item Metrics to evaluate performance between the ground truth and simulated solutions.
\end{enumerate}

This section delineates these components in addition to the tasks that comprise the benchmark and how to report and submit results. The protocols are attached and also available for download along with the dataset at: \url{https://research.csiro.au/robotics/manipulation-benchmark/}

\subsection{Tasks}

The benchmark is currently composed of 10 simple tasks which have been chosen as they provide a good initial starting point of fundamental movements and contacts. These tasks demonstrate how to use the benchmark and act as a precursor to more advanced tasks. By adapting simulators to accurately model simple tasks we theorise that these same parameters will extend to more complex scenes that share underlying correlations. The benchmark will be expanded to include more advanced tasks with a higher relevance to real-world manipulation scenarios.

The tasks are developed with the goal of being difficult in varying ways for rigid body simulators to model. That is to say, we look to exploit badly-modelled phenomena that are regularly exposed in simulation, including: (i) interaction with objects; (ii) interactions between multiple objects; (iii) dynamic movements; and (iv) manipulation of objects with contrasting inertial properties, weights, or frictions. As rigid body simulators are not designed to model deformable objects or liquids, these scenarios are omitted. Table \ref{taskTable} lists the proposed tasks together with a short description along with their containing subgroup.

\begin{table}
\vspace{4mm}
\centering
\caption{Preliminary Tasks with Short Description and Subgroup Allocations}
\label{taskTable}
\resizebox{\linewidth}{!}{%
\begin{tabular}{|l|l|l|} 
\hline
Subgroup & Task & Description \\ 
\hline
\multirow{2}{*}{Kinematics} & 1 & Kinematic motion of the arm, Simple \\ 
\cline{2-3}
 & 2 & Kinematic motion of the arm, Complex~ \\ 
\hline
\multirow{8}{*}{Non-prehensile manipulation} & 3 & Plastic cube pushing \\ 
\cline{2-3}
 & 4 & Wooden cube pushing \\ 
\cline{2-3}
 & 5 & Plastic cylinder rolling \\ 
\cline{2-3}
 & 6 & Wooden cylinder rolling \\ 
\cline{2-3}
 & 7 & Plastic cone rolling \\ 
\cline{2-3}
 & 8 & Wooden cone rolling \\ 
\cline{2-3}
 & 9 & Plastic cuboid pushing \\ 
\cline{2-3}
 & 10 & Wooden cuboid pushing \\ 
\hline
\end{tabular}
}
\end{table}

Templates based on the YCB Protocol Template \cite{Calli2015TheResearch} provide full details about the tasks, including the starting configuration and control of the robot. The positions for the robot to complete a task are given in the protocol as a series of joint rotations and corresponding times to complete each motion. 

Objects have either been 3D printed using Stratasys ASA plastic or machined from white mahogany wood. The weights for each of the objects are listed on the benchmark's website and within the task protocol. Due to the difficulty and variability inherent in measuring frictional properties \cite{Berger2002FrictionSimulation} we publicise the material but consciously neglect to include the frictional properties. We look to expand the benchmark in future to include measured frictional parameters.

% Add stuff about the objects, i.e. the material properties - plastic are made with this, wooden are made with this. We don't provide coefficients of friction because

% -Why/How we chose the tasks - Things that are tough for simulators to get right, manipulation context
% -The tasks and a short description of each
%     -Description should include why we chose it, does it compliment another test or does it provide new information

% [DON'T GO THROUGH]
% -Each task in detail, i.e. positions, setup, etc.

\subsection{Dataset}

The dataset is collected in CSIRO's Qualisys motion capture system, which we submit as a real-time high accuracy ground truth. The system comprises 24 cameras mounted on a $8\times8\times4$ meter gantry. Calibration is to a residual of $<1mm$, with the system recording at 100Hz. The latency of the system is dependant on several variables including the number of markers, number of cameras, and the computer setup.  Data is streamed from a host PC (which receives data with $<6ms$ of delay) to a 3\textsuperscript{rd} party PC running the Robotic Operating System (ROS). As entire projects have focused on measuring the latency across motion capture systems, we continue under the assumption that we receive the motion capture data in real-time and direct the reader's attention to related literature \cite{Qualysis2009Real-timeSwitzerland,Thoni2011AnalyseUmgebungen}.

The physical setup for recording the motion capture data follows the protocols set out in each task's YCB template. The base configuration for all tasks is a level laminated table ($1.8\times0.75$ meters) with a Kinova Mico\textsuperscript{2} fastened at one end. The laminated table with adjustable feet was levelled in the motion capture system using a spirit level, the table is stable and capable of supporting the 5kg Kinova arm without movement. The Kinova arm is the 6DOF spherical version attached with a KG-3 gripper, the Kinova arm was chosen as it is a common platform for the research community and although not marketed as highly repeatable it poses as a standard system used for transferring controllers across the reality gap.

\begin{figure}[tb]
\vspace{2mm}
	\centering
	\includegraphics[width=\linewidth]{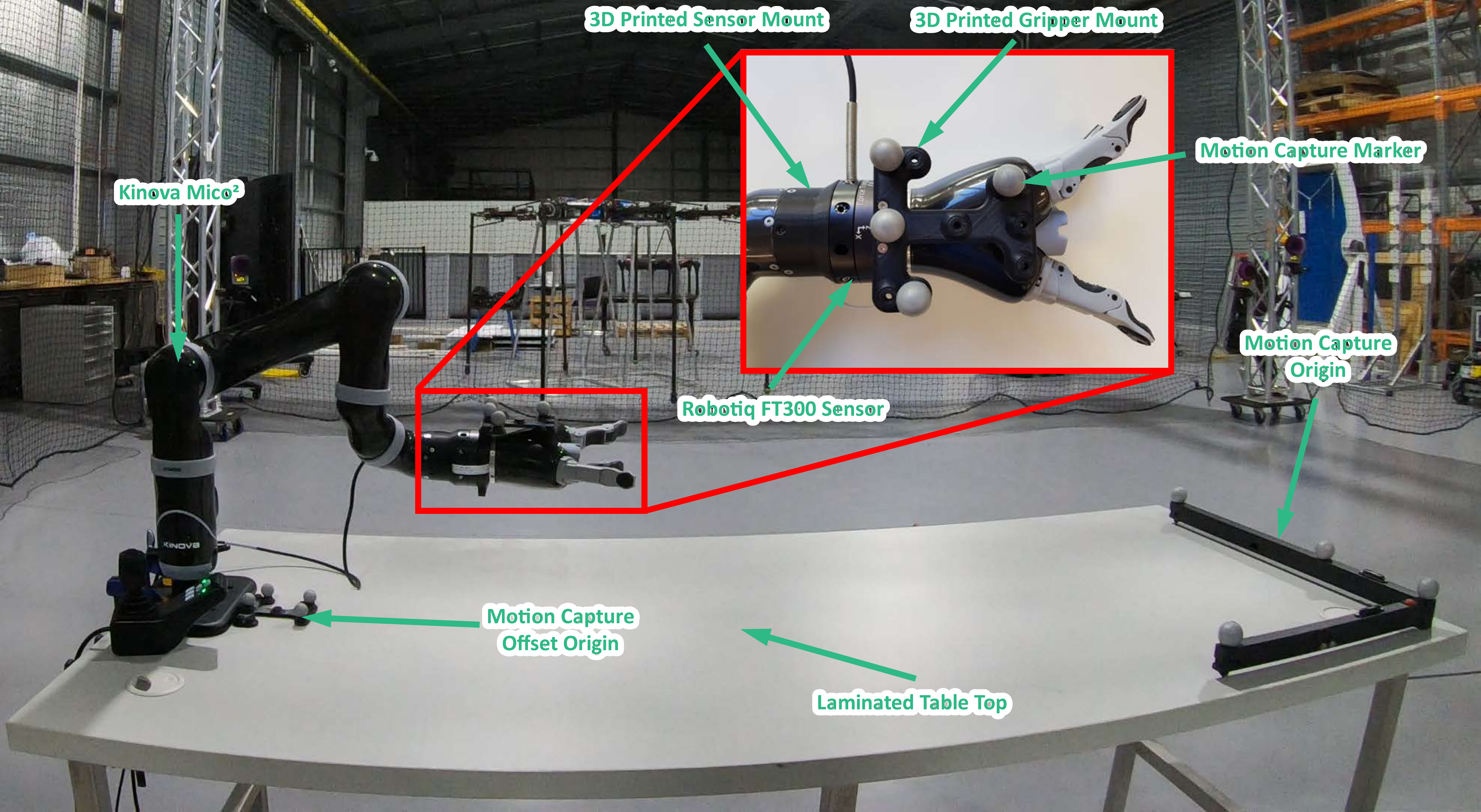}
	\caption{The base setup of the real world dataset including a magnified region of the wrist depicting the additional 3D printed mounts and the Robotiq FT3000 sensor.}
	\label{Setup}
\end{figure}

The arm is fitted with a Robotiq FT300 force torque sensor mounted between the wrist of the arm and the gripper using two 3D printed mounts as can be seen in Figure \ref{Setup} (Mesh files of the mounts can be found at the benchmarks website). The mounting brackets are 3D printed using ABS plastic which is rigid and lightweight, this satisfies the needs of the design as the final link of the system including the gripper weighs under 1.2kg. The Robotiq FT300 was calibrated with the Kinova gripper before recording of the dataset.

A proportional controller provides Velocity control of the 6 joints of the Kinova arm. The proportional controller gain is 4 with the hard limit for the joint velocities as 10 degrees/second for joints 1, 2 and 3 while joints 4, 5 and 6 have a limit of 20 degrees/second. The controller relies on the joint rotations sent from the Kinova at 10Hz, making the control cycle frequency 10Hz. The three-fingered KG-3 gripper is controlled using Kinova's position controller; the documentation\footnote{https://github.com/Kinovarobotics/kinova-ros} states encoder values for fully open and fully closed are 0 and 6400 respectively with equivalent rotations being 0 and 1.4 radians.

The software setup for recording the dataset is identical for each task. ROS is used to control the Kinova arm and to bag the received data, including: 

\begin{itemize}
    \item 6DOF pose (translation in x,y,z and rotation as a quarternion) of each tracked rigid body (Streamed at 100Hz, see the task protocol for the tracked bodies for specific tasks);
    \item Force Torque data from the Robotiq sensor mounted at the wrist (Streamed at 100Hz, force in the x, y and z axis along with moments in the x, y and z axis);
    \item Joint Torques (Streamed at 10Hz, for the 6 joints of the Kinova); and
    \item Finger Positions (Streamed at 10Hz, for the three fingers).
\end{itemize}

Each task is repeated 20 times to produce reliable measurements and create a distribution of the final static position of objects. ROS bag data is converted into a CSV file for ease of use, which we refer to as the \textit{raw data}. Each CSV file contains a metadata header, including the task number, the repeat of the task, the date and time of recording, the temperature and humidity at the time of recording and a short description of the contained information. 

% A summary 
A CSV file is generated for each task with similar header and layout. The data within the summary CSV is the mean data for the arm pose, force torque sensor, joint torques and finger positions, but includes the 6DOF pose data for objects across all 20 repeats. This data is provided at 10Hz as it is the lowest frequency of the recorded data streams. 

This dataset is provided with the intention to be built upon in future with additional tasks completed by a range of robotic manipulators. It is unnecessary for users of the benchmark to record tasks to contribute to the dataset with the intention that users apply the provided dataset to benchmark their simulation environments as detailed in \cref{Simulator Setup,Performance Metrics,Reporting Performance}.

%Must say something about 20 repeats.

% -Intro as to the system we are using
%     -Qualisys 24cameras, all experiments were recorded with a residual of less than 1mm
%     -Recordings occur at 100Hz
%     -Latency - has been reported in the past to be on the scale of milliseconds and therefore the data stream was assumed to be real time as measuring real-time latency is a project unto itself.
%     -All reported trajectories are 100\% measured without gap filling algorithms
% -Performing experiments
%     -Each of the 27 unique tasks were set up aalong with the ccording to the specifications in the YCB protocol and run 20 times each
%     -The robot was controlled through ROS with an identical control policy as described for simulation.
%     -The data was saved in QTM's .QTM file format as well as bagged with other data and provided as raw data in CSV file format at the homepage

\subsection{Simulator Setup}\label{Simulator Setup}

To benchmark a simulator using our benchmarking system there are procedures that must be followed. Any parameter that is not listed as a controlled variable is able to be exploited to improve the simulation. In general, the scene, robot and control of the robot are all set as immutable whilst most other parameters are user definable. 

Our benchmark works for any simulator. The manufacturers Unified Robot Description Format (URDF) is supplied at the benchmark's website (with alterations to account for the Robotiq FT300) as the description of the robotic manipulator along with the mesh files. If a simulator does not natively support URDF the robot is able to be assembled in the chosen simulation environment by importing the rigid body meshes following the URDF or manufacturers specifications.

The creation of the simulation scene including manipulable objects should follow the task protocol. The protocol dictates the position and orientation of the robot and objects including the reference frame used for measurement. Manipulable objects can use a simulators primitive shape functionality (if the correct shape is available) or the stereolithography file (STL) provided within the dataset. The STL can be converted into another mesh format as required.

The simulated robot is controlled using joint velocities with a proportional controller. A task includes a sequence of target joint positions with a set amount of time to achieve each movement; these movements are set out in each of the task protocols. The weighting for the proportional controller is also specified in the task protocol. The control loop is to be run at 10Hz which follows the control frequency of the dataset. 

Twenty repeats of the simulation must be recorded for comparison with the dataset. The repeats are to be saved in a single folder with iterative names for repeats (i.e. task01\textunderscore01.csv through to task01\textunderscore20.csv). Data is to be recorded at 10Hz and is to be the same as the data recorded for the dataset including: 6DOF pose of specified rigid bodies, force torque data of the wrist, joint torques for the robotic manipulator, and finger positions. The CSV must follow the ordering specified in the specific task protocol and rotations must be reported as quarternions (w,x,y,z).

\subsection{Performance Metrics} \label{Performance Metrics}

No single metric can objectively assess the performance of a simulator across all tasks, therefore a range of metrics are proposed.  In deriving suitable performance metrics, we note that averaging the trajectory of an object with the same start and end position is possible especially for an object that follows the same controlled trajectory such as the robotic manipulator. However, averaging the trajectories for objects that do not share the same end pose is not a valid computation and therefore we propose to analyse the distribution of the end pose of such objects. As there are a limited number of plausible ways of progressing from a shared start pose to an end pose whilst obeying physics, by analysing the end configuration of manipulable objects we submit that this is a valid metric. In addition, further metrics are used to characterise the movement of such objects.

We supply a script that computes the following metrics and supplies results for the benchmark. The script operates on the CSV files for the 20 simulation repeats.
% The arm has the same start and end position allowing for the mean to be calculated.

\subsubsection{Euclidean distance Error}

The Euclidean distance error is the cumulative error between positions of the real manipulator and simulated manipulator, averaged across the number of data points. This metric gives a single number description of the translational difference of the simulated and real arms. Equation \ref{euclideanError} is the implementation of the algorithm where $d_{x,y,z}$ are the mean positions of the 20 repeats of the dataset, $s_{x,y,z}$ are the mean positions of the simulator and $n_{points}$ is the number of data points.

\begin{equation}
euclidean_{error} =  \displaystyle\frac{\sum\sqrt[]{(d_{x}-s_{x})^{2}+(d_{y}-s_{y})^{2}+(d_{z}-s_{z})^{2}}}{n_{points}}
\label{euclideanError}
\end{equation}

\subsubsection{Inner product of Unit Quarternion Error}

This metric is a measure of the cumulative error of rotation for the robotic arm. The inner product of the unit quarternion for a data point is a value between $0$ and $\frac{\pi}{2}$ \cite{Huynh2009MetricsAnalysis}. In Equation \ref{rotationError}, $q_{d,s}$ are the quarternions of the dataset and simulation.

\begin{equation}
rotation_{error} =  \displaystyle\frac{\sum\arccos{(|q_d\cdot q_s|)}}{n_{points}}
\label{rotationError}
\end{equation}

\subsubsection{Pose Error}

It is possible to combine both translation and rotation into a single metric defined by a distance on the Euclidean group SE(3). Park \cite{Park1995DistanceDesign} defined one such metric as seen in Eq.~\ref{poseError}. The metric requires an appropriate length scale $r$ which can be determined using engineering considerations to equally weight both rotation and translation with knowledge on the Cartesian workspace volume. As the volume of the rotational component of Eq.~\ref{poseError} is $8\pi^2$, $r$ can be calculated as $37$ for the Cartesian volume of the Kinova arm. In addition, $\lVert \cdot \rVert$ is the Euclidean norm, $\theta_{d,s}$ are the rotation matrices for the dataset and simulation respectively and $b_{d,s}$ are the translation vector for the dataset and simulation respectively.

\begin{equation}
pose_{error} =  \displaystyle\frac{\sum\sqrt{\lVert\log(\theta_d^{-1}\theta_s)\rVert^{2}+r\lVert b_s - b_d\rVert^2}}{n_{points}}
\label{poseError}
\end{equation}

\subsubsection{Velocity Max, Average and Error}

Metrics characterised by the velocity of either the arm or objects provide insight into whether the simulation displays similar kinetic energy to that of the dataset. Velocity is derived from the change in recorded positions over time to provide an unbiased velocity for both simulation and the dataset. Maximum velocity and the velocity error between the dataset and simulation over time are calculated for the end effector. The error between dataset and simulation can not be calculated for manipulable objects, therefore the maximum velocity and average velocity over moving time are calculated.

\subsubsection{Acceleration/Deceleration Max, Average and Error}

Further characterising the motion of rigid bodies is the acceleration or deceleration of rigid bodies. By comparing the maximum acceleration and deceleration for both the arm and manipulable objects, insight is gained as to whether the simulator is mirroring the real world realistically. The average acceleration for manipulable objects over time and the acceleration error over time for the arm are also meaningful metrics. Similar to velocity, acceleration is derived from the the change in recorded position over time squared. 

\subsubsection{Motor Torque Min, Max and Error}

The torques required by the manipulator to follow the commanded trajectory show the effort necessary for the arm to operate in an environment, in this case the minimum torque is useful as it shows the energy required to oppose gravity. The minimum and maximum torques are found from the addition of the absolute torque across all 6 joints at each data point. The equation for calculating the torque error is found in Eq.~\ref{torqueError} where $\tau_{d}$ is the sum of all 6 absolute joint torques from the database and likewise $\tau_{s}$ is the simulation. 

\begin{equation}
torque_{error} =  \displaystyle\frac{\sum(\tau_{d} - \tau_{s})}{n_{points}}
\label{torqueError}
\end{equation}

\subsubsection{Contact Force and Contact Moment Max and Error}

Forces and moments measured from the wrist mounted force torque sensor provide valuable information about the contacts that the gripper makes with objects. The maximum force ($F_x+F_y+F_z$) and maximum moment ($M_x+M_y+M_z$) exerted onto the gripper at a data point make obvious whether the arm and contacting objects have the correct properties. The averaged error between the dataset and simulation across the number of data points is calculated using Eq.~\ref{forceError} for contact force and Eq.~\ref{momentError} for contact moments. Where $F_{xd,yd,zd}$ are the dataset forces and $F_{xs,ys,zs}$ are the simulation forces, similar to the moments in Eq.~\ref{momentError}.

\begin{equation}
error =  \displaystyle\frac{\sum F_{xd}+F_{yd}+F_{zd} - \sum F_{xs}+F_{ys}+F_{zs}}{n_{points}}
\label{forceError}
\end{equation}

\begin{equation}
error =  \displaystyle\frac{\sum M_{xd}+M_{yd}+M_{zd} - \sum M_{xs}+M_{ys}+M_{zsd}}{n_{points}}
\label{momentError}
\end{equation}

\subsubsection{Moving Time}

The moving time is the time in seconds from when the first rigid body moves from static until the last moving rigid body becomes static, this excludes the robotic manipulator. This metric aides in characterising the movement of objects.

\subsubsection{Distribution Comparison for Translation and Rotation}

\begin{figure*}[htb!]
    \vspace{2mm}
	\centering
	\includegraphics[width=\linewidth]{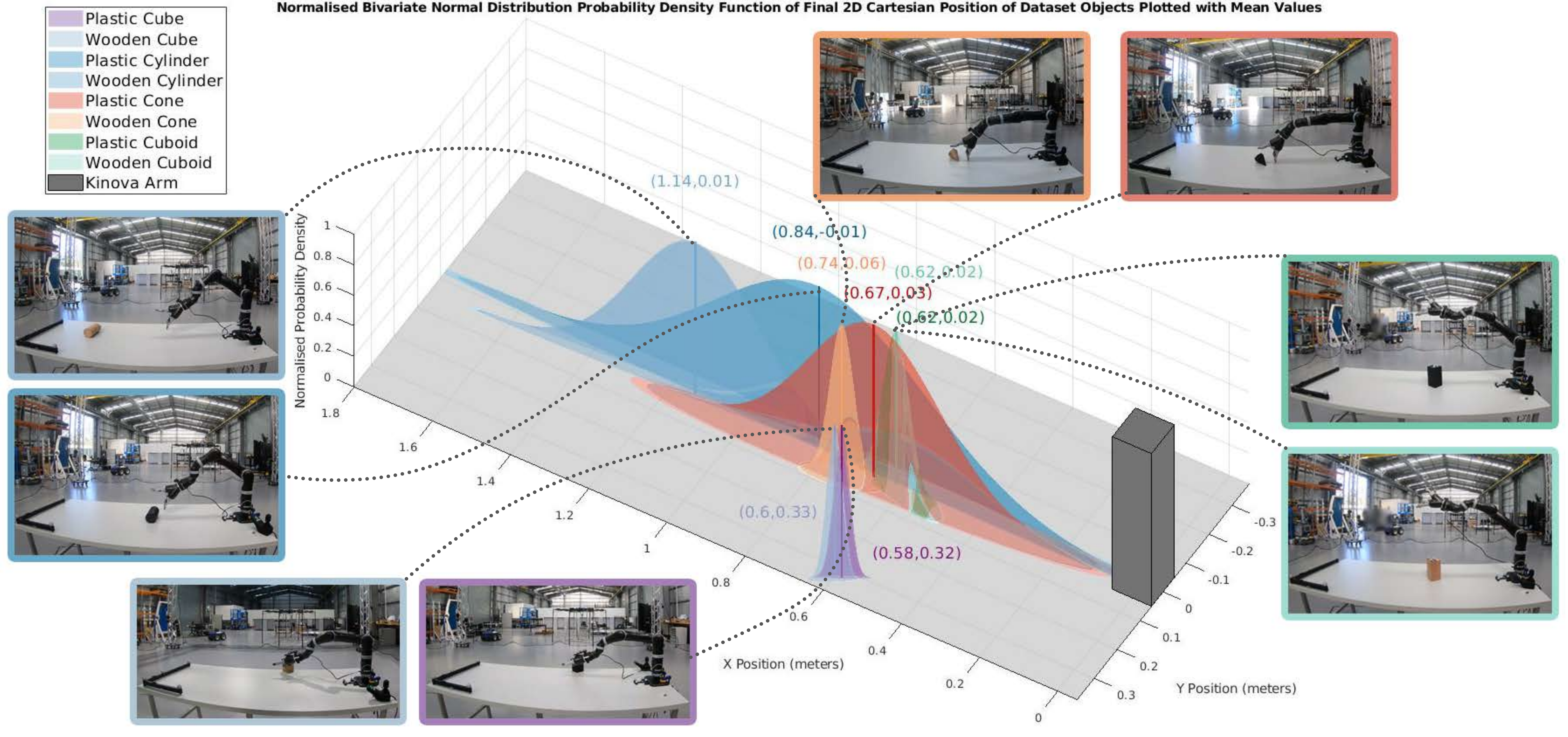}
	\caption{3D plot of the bivariate normal distribution probability density function (x and y Cartesian positions) of manipulable object final positions across 20 repeats. An image of a single repeat of the final static resting pose for tasks 3 through to 10 is included.}
	\label{TableDistribution}
\end{figure*}

As the trajectory of a manipulable object isn't directly comparable between the dataset and simulation the final resting pose is compared instead. From the dataset a multivariate normal distribution of the Cartesian position and a multivariate normal distribution of the Euler rotation is created from the 20 sample points. A visualisation of the bivariate (x and y position) distribution for the 8 tasks with manipulable objects can be seen in Fig.~\ref{TableDistribution}. The final resting Cartesian position of the manipulable object in simulation is compared to the normal distribution of the dataset using the Mahalanobis distance \cite{Mahalanobis1936OnStatistics}. The same is done for the final Euler rotation of the simulated object, resulting in a distance metric for both the translation and rotation.

\subsection{Reporting Performance} \label{Reporting Performance}

A website for the combined benchmark and dataset is intended to host the results from users of the benchmark. As there are 23 metrics for tasks including manipulable objects (15 for those without) there are too many metrics to display publicly and compare, so instead the errors will be used. Tasks are divided into subgroups based on theme, with results reported for a subgroup. Reporting results for a subgroup is seen as more targeted and efficient than having to complete all tasks, or reporting only singular tasks. 

To calculate the results for a subgroup, a mean error is taken comprising the errors in Euclidean distance, rotation, pose, velocity, acceleration, torque, force, moments, object translation and object rotation for every task in that subgroup.  Although not all the metrics are to be displayed and compared, the additional metrics will be calculated and made available as an appendix as they provide deeper insight than the error metrics alone.

To report performance of a simulator, the raw data files from the simulator should be submitted to the benchmark website using the provided form. The 20 CSV data files, in the specified layout and including metadata, should be provided for all tasks in a subgroup to be displayed on the benchmark results page. Along with the raw data files, information on the simulator environment will be required. Supplied information on the simulator should include:

\begin{itemize}
    \item Simulator used
    \item User definable parameters tuned
    \item Running time for the tasks with system specifications
\end{itemize}

%%%%%%%%%%%%%%%%%%%%%%%%%%%%%%%%%%%%%%%%%%%%%%%%%%%%%%%%%%%%%%%%%%%%%%%%%%%%%%%%
\section{Exemplar Simulator Performance}

Along with the proposal of our benchmark we submit performances from two robotic simulators. The exemplar performance is to act as a minimum benchmark with the majority of settings left to their default values. The two chosen simulators are V-Rep \cite{Rohmer2013V-REP:Framework} and PyBullet \cite{E.CoumansandY.Bai2016PybulletLearning} as they are commonly used by the robotics community. The code for running this baseline can be found at the benchmark website.  V-Rep has a range of physics engines that can be applied through an abstraction layer, therefore we are able to benchmark the following simulator and physics engine combinations:
\begin{itemize}
    \item PyBullet (Bullet 3)
    \item V-Rep (Bullet 2.78)
    \item V-Rep (Bullet 2.83)
    \item V-Rep (Newton)
    \item V-Rep (Open Dynamics Engine, ODE)
    \item V-Rep (Vortex)
\end{itemize}

\subsection{Approach}

As illustrated in Section~\ref{Simulator Setup}, both PyBullet and V-Rep are prepared for all 10 tasks according to the respective protocol. Both simulators provide native support for importing URDF files. For scenes requiring additional objects, the meshes of the objects are imported with the measurement frame positioned according to the protocol. The object is moved to the correct starting pose. 
% The starting configuration of scenes for both simulators is displayed in Figure~\ref{simSetup}.

% \begin{figure*}[tbh!]
% 	\centering
% 	\includegraphics[width=\linewidth,height=6cm]{.png}
% 	\caption{An images which is 2x5 pictures. Top line will be Pybullet scenes setup and below will be V-Rep. Maybe three lines with the real world?}
% 	\label{simSetup}
% \end{figure*}

The majority of settings for both simulators were kept to default values including: step sizes, integrator parameters, solver parameters, sensor properties, actuator properties, inertial properties, and friction models. The default frictional coefficients used by PyBullet are $0.5$ for the lateral friction and $0$ for both the rolling and spinning. V-Rep physics engines Bullet and ODE use frictional coefficients of $0.71$, Vortex uses a value of $0.5$ for the linear primary axis friction and for Newton both the static and kinetic frictions are $0.5$. Settings that were changed were limited to the position of the base of the robot arm (i.e. the arm is fixed $0.055mm$ above the surface plane) to account for the pedestal the arm is mounted on, the weight of the objects being interacted with (i.e. the cube, cylinder, cone and cuboid) which is changed to the value specified in the protocol, and for V-Rep the actuated finger joint settings were changed to “Lock motor when target velocity is zero” as torque values are not recorded for these joints.

% -Each setup follows the provided YCB protocols for each task faithfully
% -Robots are imported using the official URDF through their respective plugins, scripts, etc.
% -All settings that are not given in the protocol are left to the simulators generic values 

\subsection{Results}

\begin{table*}[htb!]
\centering
\caption{Error Metrics for Subgroup 1}
\label{subgroup1}
\resizebox{\linewidth}{!}{%
\begin{tabu}{|>{\hspace{0pt}}p{0.235\linewidth}|>{\hspace{0pt}}p{0.096\linewidth}|>{\hspace{0pt}}p{0.083\linewidth}|>{\hspace{0pt}}p{0.075\linewidth}|>{\hspace{0pt}}p{0.077\linewidth}|>{\hspace{0pt}}p{0.121\linewidth}|>{\hspace{0pt}}p{0.075\linewidth}|>{\hspace{0pt}}p{0.11\linewidth}|>{\hspace{0pt}}p{0.11\linewidth}|} 
\hline
\textbf{Simulator/Physics Engine} & \textbf{Euclidean} & \textbf{Rotation} & \textbf{Pose} & \textbf{Velocity} & \textbf{Acceleration} & \textbf{Torque} & \textbf{Force} & \textbf{Moment} \\ 
\hline
Pybullet & 0.1311 & 1.1049 & 15.4065 & 0.1487 & 0.6156 & 36.6548 & 10.5721 & 0.1211 \\ 
\hline
V-Rep Bullet2.78 & 0.1990 & 1.0861 & 17.2124 & 0.1425 & 0.6479 & 3.7073 & 6.9713E+07 & 6.9707E+07 \\ 
\hline
V-Rep Bullet2.83 & 0.2131 & 1.0850 & 15.3668 & 0.1428 & 0.6775 & 4.1507 & 6.9713E+07 & 6.9707E+07 \\ 
\hline
V-Rep Newton & 0.1292 & 1.1216 & 25.0602 & 0.1516 & 0.6170 & 4.0919 & 6.9713E+07 & 6.9707E+07 \\ 
\hline
V-Rep ODE & 0.1546 & 1.1154 & 17.9188 & 0.1552 & 0.6275 & 3.8730 & 6.9713E+07 & 6.9707E+07 \\ 
\hline
V-Rep Vortex & 0.1288 & 1.1216 & 24.8870 & 0.1518 & 0.6276 & 4.0782 & 6.9713E+07 & 6.9707E+07 \\
\hline
\end{tabu}
}
\end{table*}

% \begin{table*}[htb!]
% \centering
% \caption{Error Metrics for Subgroup 2}
% \label{subgroup2}
% \resizebox{\linewidth}{!}{%
% \begin{tabu}{|>{\hspace{0pt}}p{0.177\linewidth}|>{\hspace{0pt}}p{0.071\linewidth}|>{\hspace{0pt}}p{0.062\linewidth}|>{\hspace{0pt}}p{0.056\linewidth}|>{\hspace{0pt}}p{0.058\linewidth}|>{\hspace{0pt}}p{0.091\linewidth}|>{\hspace{0pt}}p{0.056\linewidth}|>{\hspace{0pt}}p{0.081\linewidth}|>{\hspace{0pt}}p{0.083\linewidth}|>{\hspace{0pt}}p{0.131\linewidth}|>{\hspace{0pt}}p{0.114\linewidth}|} 
% \hline
% \textbf{Simulator/Physics Engine} & \textbf{Euclidean} & \textbf{Rotation} & \textbf{Pose} & \textbf{Velocity} & \textbf{Acceleration} & \textbf{Torque} & \textbf{Force} & \textbf{Moment} & \textbf{Object Translation} & \textbf{Object Rotation} \\ 
% \hline
% Pybullet & 0.0727 & 1.1759 & 10.0995 & 0.0561 & 0.2725 & 40.0083 & 10.6550 & 0.0960 & 31421.8032 & 197.4230 \\ 
% \hline
% V-Rep Bullet2.78 & 0.0780 & 1.1592 & 10.1288 & 0.0587 & 0.2887 & 2.0007 & 3.9844E+07 & 1.6920E+29 & 38.6421 & 428.7655 \\ 
% \hline
% V-Rep Bullet2.83 & 0.0814 & 1.1535 & 9.9682 & 0.0652 & 0.3735 & 2.3743 & 4.0235E+07 & 1.3536E+29 & 53.2329 & 257.1772 \\ 
% \hline
% V-Rep Newton & 0.0724 & 1.1834 & 11.2525 & 0.0622 & 0.2726 & 1.9361 & 4.3754E+07 & 4.3750E+07 & 46.2465 & 209.3794 \\ 
% \hline
% V-Rep ODE & 0.0665 & 1.1686 & 10.1306 & 0.0629 & 0.2824 & 2.0974 & 4.0235E+07 & 1.5228E+29 & 34.0668 & 99.8609 \\ 
% \hline
% V-Rep Vortex & 0.0717 & 1.1830 & 11.0406 & 0.0623 & 0.2747 & 1.9407 & 3.9844E+07 & 1.6920E+29 & 85.8384 & 203.6449 \\
% \hline
% \end{tabu}
% }
% \end{table*}

\begin{table*}[htb!]
\centering
\caption{Error Metrics for Subgroup 2}
\label{subgroup2}
\resizebox{\linewidth}{!}{%
\begin{tabu}{|>{\hspace{0pt}}p{0.177\linewidth}|>{\hspace{0pt}}p{0.071\linewidth}|>{\hspace{0pt}}p{0.062\linewidth}|>{\hspace{0pt}}p{0.056\linewidth}|>{\hspace{0pt}}p{0.058\linewidth}|>{\hspace{0pt}}p{0.091\linewidth}|>{\hspace{0pt}}p{0.052\linewidth}|>{\hspace{0pt}}p{0.083\linewidth}|>{\hspace{0pt}}p{0.083\linewidth}|>{\hspace{0pt}}p{0.131\linewidth}|>{\hspace{0pt}}p{0.116\linewidth}|} 
\hline
 \textbf{Simulator/Physics Engine}  & \textbf{Euclidean}  & \textbf{Rotation}  & \textbf{Pose}  & \textbf{Velocity}  & \textbf{Acceleration}  & \textbf{Torque}  & \textbf{Force}  & \textbf{Moment}  & \textbf{Object Translation}  & \textbf{Object Rotation}  \\ 
\hline
Pybullet & 0.0725 & 1.2161 & 13.1405 & 0.0562 & 0.2723 & 39.957 & 10.6504 & 0.0960 & 31467.2361 & 194.0998 \\ 
\hline
V-Rep Bullet2.78 & 0.0780 & 1.1592 & 10.1288 & 0.0587 & 0.2887 & 2.0007 & 3.9844E+07 & 1.6920E+29 & 38.6421 & 428.7655 \\ 
\hline
V-Rep Bullet2.83 & 0.0814 & 1.1535 & 9.9682 & 0.0652 & 0.3735 & 2.3743 & 4.0235E+07 & 1.3536E+29 & 53.2329 & 257.1772 \\ 
\hline
V-Rep Newton & 0.0724 & 1.1834 & 11.2525 & 0.0622 & 0.2726 & 1.9361 & 4.3754E+07 & 4.3750E+07 & 46.2465 & 209.3794 \\ 
\hline
V-Rep ODE & 0.0665 & 1.1686 & 10.1306 & 0.0629 & 0.2824 & 2.0974 & 4.0235E+07 & 1.5228E+29 & 34.0668 & 99.8609 \\ 
\hline
V-Rep Vortex & 0.0717 & 1.1830 & 11.0406 & 0.0623 & 0.2747 & 1.9407 & 3.9844E+07 & 1.6920E+29 & 85.8384 & 203.6449 \\
\hline
\end{tabu}
}
\end{table*}

%Talk about table 1
We start by noting that results are presented to establish the importance of the chosen metrics, not to rank the performance of the simulators. Table~\ref{subgroup1} displays the results for subgroup 1 (Tasks 1 and 2) in the format to be displayed on the benchmarks website. From this table we note that (i) the V-Rep simulator and physics engines have a much lower joint torque error than PyBullet, and (ii) the force torque readings from V-Rep are exponentially large when compared to PyBullet. Further analysis of the V-Rep simulations reveals this to be due to deficiency in constraint stability of the connecting wrist joint, which could be rectified with smaller time steps or an alteration of the integrator parameters. Also evident from Table~\ref{subgroup1} is the small window of deviation that the Euclidean error and rotation error display. The pose error provides a better metric in this case to differentiate between the errors in translation and/or rotation.

%Talk about table 2
Table~\ref{subgroup2} contains the results for subgroup 2 (Tasks 3 through 10) and includes two additional columns for the translation and rotation discrepancies of the manipulable object. Similar to Table~\ref{subgroup1} the torque error for Pybullet is much larger than the group of V-Rep physics engines, contrasting the large discrepancies in force and moment measurements recorded by V-Rep. Pose error again remains a valuable and informative metric. In terms of the errors associated with the manipulable object V-Rep ODE is closest in both translation and rotation (Translation: $34.07$, Rotation: $99.86$). Conversely to the low error of ODE, Pybullet has a large translational error of $31467.2$ which is caused by a considerable discrepancy in the cylinder task. 

%Talk about additional metrics
Benchmark metrics not reported in the table permit further insight into some of the great disparities in the error metrics, and also how improvement in the environment could be achieved. Comparing the maximum velocities of the arm between all simulators and tasks the maximum was in the range of $0.32$ and $0.69 m/s$ whereas the dataset has a maximum of between $0.56$ and $1.36 m/s$ across all tasks, almost double that of the simulators. The same is evident in the acceleration of the arm, simulators were between $0.89$ and $2.44 m/s^2$ while the arm in the real world experienced accelerations of between $1.01$ and $3.39 m/s^2$. This indicates that the parameters that control the behaviour of the arm in simulation need to be tuned to attain higher velocities and greater accelerations. 

The moving time of objects in simulation is also of importance as this metric assists in characterising the movement of the object and as a result the error in the final pose. The greatest disparity between the simulators and the real world are in the rolling tasks. The rolling time of the dataset cylinder was $68.31\pm7.01$ s for plastic and $64.32\pm12.1$ s for wood. PyBullet was the closest simulator ($75.8$ seconds for plastic and $75.7$ seconds for wood). In the V-Rep engines, no end effectors made contact with the cylinder during the predefined control sequence for tasks 5 and 6 and therefore resulted in a stationary cylinder. This lack of contact is likely due to the discrepancies between actuator response in the real world and simulation, highlighting the need for the physics engine parameters to be tuned and displaying the usefulness of the moving time metric.

The cone task saw similar results, with a rolling time of $58.09\pm3.86$ seconds for plastic and $55.7\pm1.73$ seconds for wood.  In this case PyBullet achieved the least accurate result ($6.5$ seconds for plastic and $6.5$ seconds for wood) with ODE closest to replicating the rolling time of the dataset distribution ($68.26$ seconds for plastic and $67.7$ seconds for wood). 

% All sims maximum velocities for arm between 0.32m/s and 0.69. Dataset has a maximum of between 0.56 and 1.36 dependant on task
% All sims maximum acceleration for arm between 0.89m/s^2 and 2.44. Dataset has a maximum of between 1.01 and 3.39 dependant on task
% All sims maximum deceleration for arm between 0.45m/s^2 and 5.63. Dataset has a maximum of between 2.37 and 13.62 dependant on task

% Maximum force applied from dataset = 3.27-7.28
% Maximum force by PyBullet ~16N for all tasks

% moving time for cyclinder and std dev
% dataset: plastic=68.305+-7.01, wooden=64.32+-12.1
% closest sim: P=75.9, W=75.7 PyBullet
% worst sim: All v-rep as there is no contact with cylinder

% moving time for cone
% dataset: plastic=58.09+-3.86, wooden=55.7+-1.73
% closest sim: ODE P=68.26, W=67.7 followed by bullet2.83
% worst sim:Pybullet P=6.2, W= 6.39

%%%%%%%%%%%%%%%%%%%%%%%%%%%%%%%%%%%%%%%%%%%%%%%%%%%%%%%%%%%%%%%%%%%%%%%%%%%%%%%%
\section{Conclusion}

In conclusion, our benchmark for simulated robotic manipulation validates simulation environments by drawing comparisons via metrics between simulation and a ground truth dataset. There are 23 metrics presented that comprehensively characterise the discrepancies caused by the reality gap and assist in benchmarking the results of simulation environments. The dataset supplies valuable information including 6DOF pose from motion capture, joint torques, and forces and moments experienced at the wrist of the robotic manipulator.

We also compare the performances of two common robotic simulators, V-Rep and PyBullet, with generic parameter settings. The precision of the simulators when completing the dataset tasks is analysed using our submitted metrics, demonstrating the practicality of the chosen metrics. We expect users of the benchmark to improve on the reported results.

Our benchmark and dataset are designed to assist researchers and developers in quantifying the reality gap, specifically for physical interaction with robotic manipulators. All necessary materials to utilise the benchmark including dataset, videos, protocols and exemplar simulator code can be found on the benchmark website. In future we look to expand the dataset to include further tasks and manipulators.

%%%%%%%%%%%%%%%%%%%%%%%%%%%%%%%%%%%%%%%%%%%%%%%%%%%%%%%%%%%%%%%%%%%%%%%%%%%%%%%%

%%%%%%%%%%%%%%%%%%%%%%%%%%%%%%%%%%%%%%%%%%%%%%%%%%%%%%%%%%%%%%%%%%%%%%%%%%%%%%%%

\addtolength{\textheight}{-11cm} 



% \section*{ACKNOWLEDGMENT}

% The preferred spelling of the word ÒacknowledgmentÓ in America is without an ÒeÓ after the ÒgÓ. Avoid the stilted expression, ÒOne of us (R. B. G.) thanks . . .Ó  Instead, try ÒR. B. G. thanksÓ. Put sponsor acknowledgments in the unnumbered footnote on the first page.

%%%%%%%%%%%%%%%%%%%%%%%%%%%%%%%%%%%%%%%%%%%%%%%%%%%%%%%%%%%%%%%%%%%%%%%%%%%%%%%%

\bibliographystyle{IEEEtran}
\bibliography{IEEEabrv,Mendeley.bib}

%                                   % the textheight of the last page by a suitable amount.
%                                   % This command does not take effect until the next page
%                                   % so it should come on the page before the last. Make
%                                   % sure that you do not shorten the textheight too much.

\end{document}